\documentclass[journal]{IEEEtran}
\IEEEoverridecommandlockouts
\usepackage{cite}
\usepackage{amsmath,amssymb,amsfonts}
\usepackage{algorithmic}
\usepackage{graphicx}
\usepackage{textcomp}
\usepackage{caption}
\usepackage{subcaption}
\usepackage{comment}
\usepackage{tabularx}
\usepackage{nomencl}
\usepackage{gensymb}
\usepackage{wrapfig}
\title{Deterioration Prediction using Time-Series of Three Vital Signs and Current Clinical Features Amongst COVID-19 Patients}

\author{Sarmad Mehrdad, \IEEEmembership{Student Member, IEEE},\\ Farah E. Shamout, Yao Wang, \IEEEmembership{Fellow, IEEE}, and S. Farokh Atashzar$^*$, \IEEEmembership{Senior Member, IEEE}
\thanks{
Sarmad Mehrdad is with the Department of Electrical and Computer Engineering, New York University (NYU). Yao Wang is with the Department of Electrical and Computer Engineering at NYU. She is also with the Department of Biomedical Engineering. Farah Shamout is with the Division of Engineering a NYU Abu Dhabi. She is also with Computer Science  and Engineering and Biomedical Engineering at NYU Tandon. S. Farokh Atashzar is with the Department of Electrical and Computer Engineering at NYU. He is also with the Department of Mechanical and Aerospace Engineering, NYU.\\%
{$^*$} Corresponding Author: S. F. Atashzar 
    {\tt\small (f.atashzar@nyu.edu)}} 
    }

\begin{document}

	\maketitle
\makenomenclature
\thispagestyle{plain}
\pagestyle{plain}
\bstctlcite{IEEEexample:BSTcontrol}

\begin{abstract}
Unrecognized patient deterioration can lead to high morbidity and mortality. Most existing deterioration prediction  models require a large number of clinical information, typically collected in hospital settings, such as medical images or comprehensive laboratory tests. This is infeasible for telehealth solutions and highlights a gap in deterioration prediction models that are based on minimal data, which can be recorded at a large scale in any clinic, nursing home, or even at the patient's home. In this study, we propose and develop a prognostic model that predicts if a patient will experience deterioration in the forthcoming 3-24 hours. The model sequentially processes routine triadic vital signs: (a) oxygen saturation, (b) heart rate, and (c) temperature. The model is also provided with basic patient information, including sex, age, vaccination status, vaccination date, and status of obesity, hypertension, or diabetes. We train and evaluate the model using data collected from 37,006 COVID-19 patients at NYU Langone Health in New York, USA. The model achieves an area under the receiver operating characteristic curve (AUROC) of 0.808-0.880 for 3-24 hour deterioration prediction. We also conduct occlusion experiments to evaluate the importance of each input feature, where the results reveal the significance of continuously monitoring the variations of the vital signs. Our results show the prospect of  accurate deterioration forecast using a minimum feature set that can be relatively easily  obtained using wearable devices and self-reported patient information.
\end{abstract}


\renewcommand{\nomname}{Nomenclature}

\nomenclature{\(LSTM\)}{Long Short-Term Memory}
\nomenclature{\(CNN\)}{Convolutional Neural Network}
\nomenclature{\(VAE\)}{Variational Auto-Encoder}
\nomenclature{\(GBM\)}{Gradiant Boosting Machine}
\nomenclature{\(CXR\)}{Chest X-Ray}
\nomenclature{\(CT\)}{Computed Tomography Scan}
\nomenclature{\(SEQ\)}{Sequential}
\nomenclature{\(EHR\)}{Electronic Health Record}
\nomenclature{\(AUROC\)}{Area Under the Receiver Operating Curve}
\nomenclature{\(AUPRC\)}{Area Under the Precision Recall Curve}
\nomenclature{\(AI\)}{Artificial Intelligence}
\nomenclature{\(MLP\)}{Multi Layer Perceptron}
\nomenclature{\(DT\)}{Decision Tree}
\nomenclature{\(CCC\)}{Clinical and Comorbidity Characteristics}
\nomenclature{\(DL\)}{Deep Learning}
\nomenclature{\(ML\)}{Machine Learning}
\nomenclature{\(FC\)}{Fully Connected}
\nomenclature{\(HR\)}{Heart Rate}
\nomenclature{\(SpO2\)}{Oxygen Saturation}

\printnomenclature

\section{Introduction}
\label{sec:introduction}
\IEEEPARstart{T}{he} significant shock imposed by the novel coronavirus (COVID-19) pandemic fundamentally challenged the delivery and management of health care services globally \cite{56}. According to the World Health Organization, more than 620 million patients have been diagnosed with COVID-19 as of October 2022, and there are around 6.52 million deaths \cite{WHO}. Since March 2020, 96.2 million patients have been admitted to emergency departments across the United States \cite{73}.

Patients with COVID-19 can experience rapid deterioration entailing the need for invasive measures associated with high morbidity or mortality \cite{57}. During the pandemic, patient prognosis was challenging, especially in the early days when the knowledge about the disease was limited, and any modifications in admission protocols could significantly alter the patient outcomes \cite{59}. This highlighted the importance of routine patient monitoring to ensure that patients with the highest risk of deterioration receive early attention \cite{58}. 

Due to the saturation of healthcare systems and concerns over unnecessary exposure, many outpatients or those in nursing centers were advised to monitor symptoms remotely and report through telemedicine \cite{63}. Hence, patients would avoid visiting emergency care facilities unless symptoms were considered significantly severe and require immediate and specialized attention \cite{64}. Although this practice could reduce exposure and unnecessary loads on emergency services \cite{65}, it could also result in poor patient prognosis. In fact, for some patients, especially those with comorbidities, the development of symptoms was followed by sudden, drastic, and unexpected deterioration resulting in morbidity, even after discharge from a clinic \cite{66}.

Considering the scale of data gathered from a plethora of patients with COVID-19 admitted to emergency departments worldwide, many DL and ML methods were developed for early diagnosis \cite{60}, patient severity assessment \cite{61}, or prognosis prediction \cite{62, 67}. In Table~\ref{tab_1}, we summarize relevant work based on choice of datasets and models for the different prediction tasks. Due to the large volume of research, it would not be possible to cite all relevant papers, hence Table I provides a balanced list.

While most of the existing work focuses on the diagnosis of COVID-19 rather than patient prognosis, many studies heavily rely on large input feature sets, specifically high-dimensional imaging, such as chest CT or X-ray scans, and other non-imaging modalities, such as laboratory test results. In addition, many of the existing models do not exploit variations in data over time. Even though the use of such data for computational models has shown great potential, there is a lack of seamlessly-scalable models based on minimal feature sets collected over time. We specifically prioritize data that can be collected not only in hospitals, but also in nursing centers or patient homes, such as using wearable devices e.g., smartwatches \cite{72}.


\begin{table}[t!]
   \centering
    \caption{\textbf{Summary of related work.} Overview of related work on the diagnosis of patients with COVID-19, patient severity assessment, and patient prognosis.}
    \begin{tabularx}{\linewidth}{|>{\hsize=0.25\hsize}X |
                              >{\hsize=0.6\hsize}X | 
                              >{\hsize=0.8\hsize}X|
                              >{\hsize=1.35\hsize}X|}
        \hline
       \textbf {Paper}  & \textbf{Task} & \textbf{Data} & \textbf{Machine learning model}\\
         \hline

         \cite{2,3,6,11,12,20,21,22,23,24,25,26,27,28,29,30,31,32,33,34,35,36} & Diagnosis & CXR, CT Scan, EHR, Clinical and comorbidities & CNN, GBM, VGG19, APACHE, ResNet50, VAE, LSTM \\ \hline
         \cite{37,38,39,40,41,48} & Severity assessment & CXR, Clinical and comorbidity features & SVM, DenseNet, CNN, MLP \\ \hline
         \cite{5,13,15,16,18,47} & Prognosis & CXR, CT Scan & CNN, DenseNet121-FPN, GLM, GBM, XGBoost, DT, AlexNet, Inception-V4 \\ \hline
    \end{tabularx}
    \label{tab_1}
    \vspace{-3mm}
\end{table}

In this study, we propose a deep neural network to model time-series of three vital signs only to predict the deterioration amongst patients with COVID-19. The ultimate goal of this work is to provide a light and scalable prediction model to support clinical decision-making for a wide range of patients on the long term, including at home patients, outpatients, and inpatients. To minimize the size of the input feature space, we focus on three basic vital signs (i.e., SpO2, HR, and temperature). This design choice  is motivated by the wide availability of wearable devices, such as smart watches, that  can monitor these vital signs. We specifically exclude other vital signs, such as blood pressure, because it cannot be measured using readily available wearable systems.

To develop and evaluate this model, we use real-world data collected at NYU Langone Health between January 2020 and September 2022. The model predicts deterioration at time horizons of 3 to 24 hours using the vital-signs time-series data collected in the 24 hours preceding the time of prediction (corresponding to the beginning of the prediction horizon), defined as in-hospital mortality, admission to the intensive care unit (ICU), or intubation. We refer to the vital-sign data as SEQ data in the remainder of this paper. The model is also provided with a small set of features reflecting CCC, including sex, age,  vaccination status, vaccination date, and status of obesity, hypertension, and diabetes (referred to as non-SEQ data).  The model includes a LSTM network \cite{68} to process the SEQ data, and a MLP that combines the last hidden state of the LSTM and the non-SEQ data. The LSTM network utilizes   temporal dilation to enable  access to longer memory dynamics without exponentially increasing the size and complexity of the computational framework. For each prediction horizon, a separate model is trained and is optimized using the commonly-used cross entropy loss through a three-phase training procedure.

\begin{table}[t]
    \centering
    \caption{\textbf{Overview of patient cohort.} We summarize in this table the patient characteristics, including demographics, and distribution of vital signs, for patients who deteriorated and patients who did not deteriorate.}
    \begin{tabularx}{\linewidth}{|>{\hsize=1.15\hsize}X |
                              >{\hsize=0.4\hsize}X | 
                              >{\hsize=0.45\hsize}X|}
        \hline
        \textbf{Characteristics} & \textbf{Deterioration} & \textbf{No Deterioration} \\
        \hline
        \textbf{Patient}, n & \textbf{6104} & \textbf{30902} \\
        \hspace{0 mm}Age (years), mean (SD) & 66.0 (18.4) & 63.1 (18.0) \\
        \hspace{0 mm}Sex (females), n (\%) & 2539 (41.5) & 17792 (57.5) \\
        \hspace{0 mm}Diabetes w/o Complications, n (\%) & 1510 (24.7) & 5582 (18.0) \\
        \hspace{0 mm}Diabetes w/ Complications, n (\%) & 114 (1.8) & 401 (1.2) \\
        \hspace{0 mm}Hypertension, n (\%) & 2721 (44.5) & 11748 (38.0) \\
        \hspace{0 mm}Vaccination n (\%) & 2388 (39.1) & 16776 (54.2) \\
        \hspace{0 mm}Time since last vacination (months), mean (SD) & -0.67 (6.2) & -0.94 (7.3) \\
        \hspace{0 mm}Obesity, n (\%) & 1003 (16.4) & 5370 (17.3) \\
        \textbf{Vital signs feature sets}&   &   \\
        \hspace{0 mm}SpO2 (\%), mean (SD) & 95.2 (4.5) & 96.2 (2.7) \\
        \hspace{0 mm}HR (Beats per minute), mean (SD) & 93.98 (27.1) & 82.9 (18.57) \\
        \hspace{0 mm}Temperature (\degree F), mean (SD) & 98.37 (1.6) & 98.2 (1.4) \\
        \hline
        
    \end{tabularx}
    \label{tab_2}
    \vspace{-3mm}
   
\end{table}

The results show that the proposed model achieves an AUROC of 0.808-0.880 in the 3 to 24 hours prediction time horizons, in a three-fold cross-validation setting. While the results are not directly comparable to those in existing work due to differences in data pre-processing, the model achieves a comparable performance. For example, the model in \cite{10}, which uses CXR images and other clinical variables achieves 0.765 AUROC in predicting deterioration within 24 hours.

In order to assess the significance of the various CCC features, and the importance of the temporal history of vital signs, we also perform a sensitivity analysis through an occlusion experiment. Overall, our work highlights the feasibility of achieving high model performance for deterioration prediction amongst patients with COVID-19 using minimal feature sets, which are easy to obtain not only in the hospital setting, but potentially also in nursing centers and at patient homes.

\section{Methodology}

\subsection{Dataset}
In this study, we use the NYU Langone De-identified COVID-19 dataset \cite{Dataset} collected from patients at the NYU Langone facilities between January 2020 and September 2022. We define an inclusion and exclusion criteria. First, in the case of multiple patient encounters, we use the patient's most recent encounter. Then, we include patients who either tested positive for COVID-19 at the facility, or were already diagnosed with COVID-19 at the time of their admission.  Next, we include in-patients with vital sign-measurements. The vital signs of these patients, including SpO2, temperature, and HR,  are periodically measured and recorded roughly every 4-5 hours. For each patient, age, sex, vaccination status and time, and the presence of comorbidities including obesity, diabetes, and hypertension are also recorded. 

Similar to previous work (see \cite{10} and references therein),  we define deterioration as the occurrence of the composite outcome of mortality, ICU admission, or intubation, i.e., any of the three events. In patient encounters with several adverse events, we only consider the occurrence of the earliest deterioration event.
It should be noted that if there are multiple deteriorations of the same type (e.g. ICU admission) recorded for a patient for more than a week apart, we only consider the latest as the reference time of deterioration for the patient.
For patients who had deteriorated, we extract vital-sign data in the 48 hours preceding the time of deterioration. 

We use this data to define ``positive'' windows for each prediction horizon, where $t=0$ represents the end of the window and $t=-24$ represents the start of the window, such that for example, in the prediction horizon of 24 hours, deterioration would have occurred at $t=24$.
For patients who did not experience deterioration and were discharged, we use the 48 hours window preceding the last vital-sign recording, and similarly use those to formulate the ``negative'' windows. We exclude all samples containing less than 48 hours of  vital-sign monitoring, either preceding the deterioration time or discharge time. 

\begin{figure}[t]
    \includegraphics[width=0.5\textwidth]{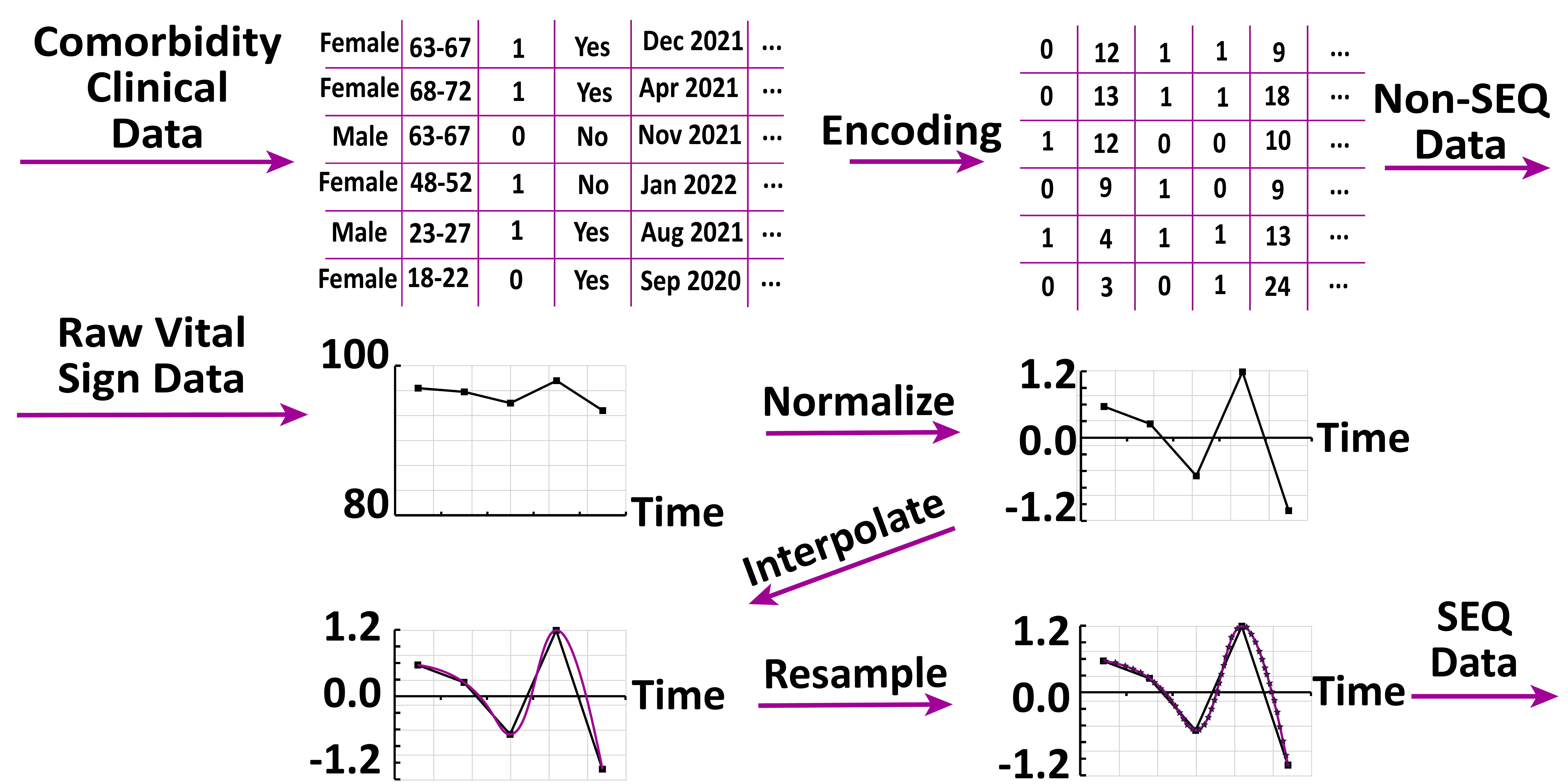}
    \vspace{-0.3cm}
    \caption{\textbf{Data pre-processing pipeline.} We encode the non-SEQ data and pre-process the SEQ data: (i) normalize via Z-score normalization, (ii) model the time-series using cubic spline interpolation, (iii) and resample at every 15 minutes.}
    \label{fig_1_preprocess} 
    \vspace{-5mm}
\end{figure}

To pre-process the time-series data, we first normalize the data using Z-score normalization based on the mean and standard deviation of each vital sign.  Since the vital signs are measured at irregular intervals, we resample each time-series to obtain regularly sampled input for the LSTM network. This resampling is done by first interpolating the raw non-uniformly sampled data through cubic spline interpolation, and then sampling the interpolated signal at every 15 minutes. In Figure~\ref{fig_1_preprocess}, we show a schematic summarizing the pre-processing of the raw time-series data, which we refer to as the SEQ data.  

As for the non-SEQ data, we encode patient sex, vaccination status, hypertension status and obesity status as binary  ($0$ or $1$). For diabetes status, we use one-hot encoding to represent if a patient is non-diabetic ($[1, 0, 0]$), diabetic without complications ($[0, 1, 0]$), or diabetic with complications ($[0, 0, 1]$). We grouped the age into $18$ different sub-groups, and replaced each age with the corresponding age sub-group (value between $1$ to $18$). For vaccination time, we count the number of elapsed months between the time of the second COVID-19 vaccination shot and the day of the time of prediction ($t=0$).

\begin{figure}[t!]
\begin{center}
    \includegraphics[width=0.5\textwidth]{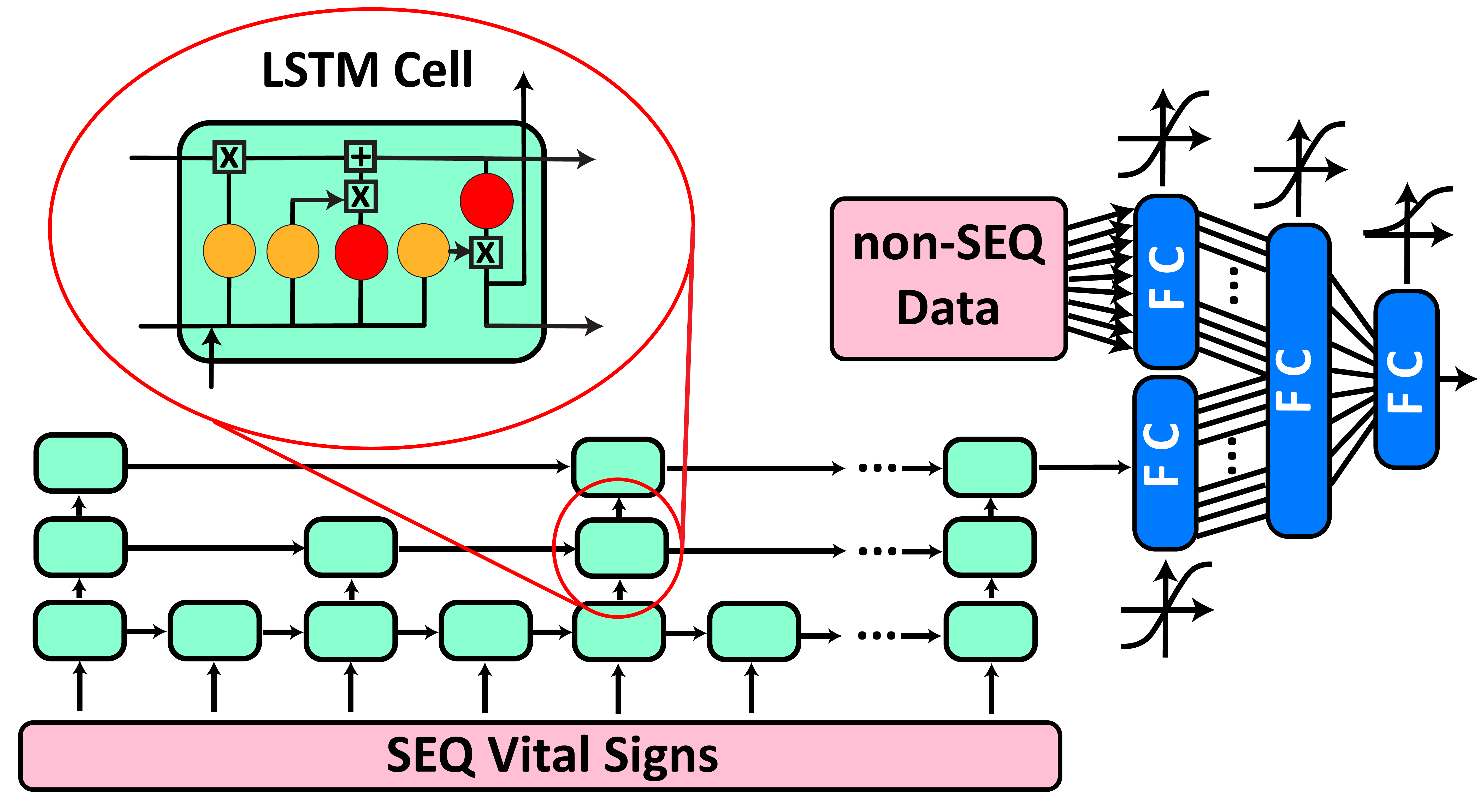}
    \end{center}
    \caption{\textbf{Architecture of the deep neural network.} The SEQ vital sign network processes the SEQ data through an LSTM-based network, whereas the non-SEQ module encompasses the non-SEQ data. 
    }
    \label{fig_4}
    \vspace{-4mm}
\end{figure}

\subsection{DL-based Deterioration Prediction Model}
Our proposed deep neural network architecture consists of LSTM layers and fully-connected (FC) layers. The overall architecture of this network is shown in Figure~\ref{fig_4} and we refer to it as the \textit{Sequential Vital Sign Network (SVS-Net)}, consisting of two modules. The SEQ data is processed by a module consisting of an LSTM network and a single FC layer, while the non-SEQ data is processed by a second module consisting of an independent FC layer. The final prediction is based on both modalities.

\vspace{2mm}
\subsubsection{SVS-Net architecture details}
LSTM networks are well-known for their ability to learn from SEQ data and have been widely used in studies where the time-series data is integral to the learning of the system and predicting future events \cite{4,71}. The temporal module consists of a temporally-dilated LSTM, which takes three vital signs as input at each time step, and has three layers each containing 32 hidden units. The final hidden state of the LSTM network, consisting of dimensionality of 32, is processed by a FC layer with an output dimensionality of 16. The non-SEQ data, dimensionality of 9, is processed by a single FC layer, which computes an output of dimensionality of 16. 

Finally, the latent representations of the two modalities are concatenated as a vector of dimensionality of 32 and then processed by a FC fusion layer with output dimensionality of 8. This is then followed by a single FC layer with sigmoid activation and an output dimensionality of one, which represents the prediction that a sample precedes deterioration or not within the specified time-horizon. All of the FC layers use hyperbolic tangent activation except for the last layer which uses sigmoid activation.

\begin{figure*}[t!]
\centering

     \includegraphics[width=0.9\textwidth]{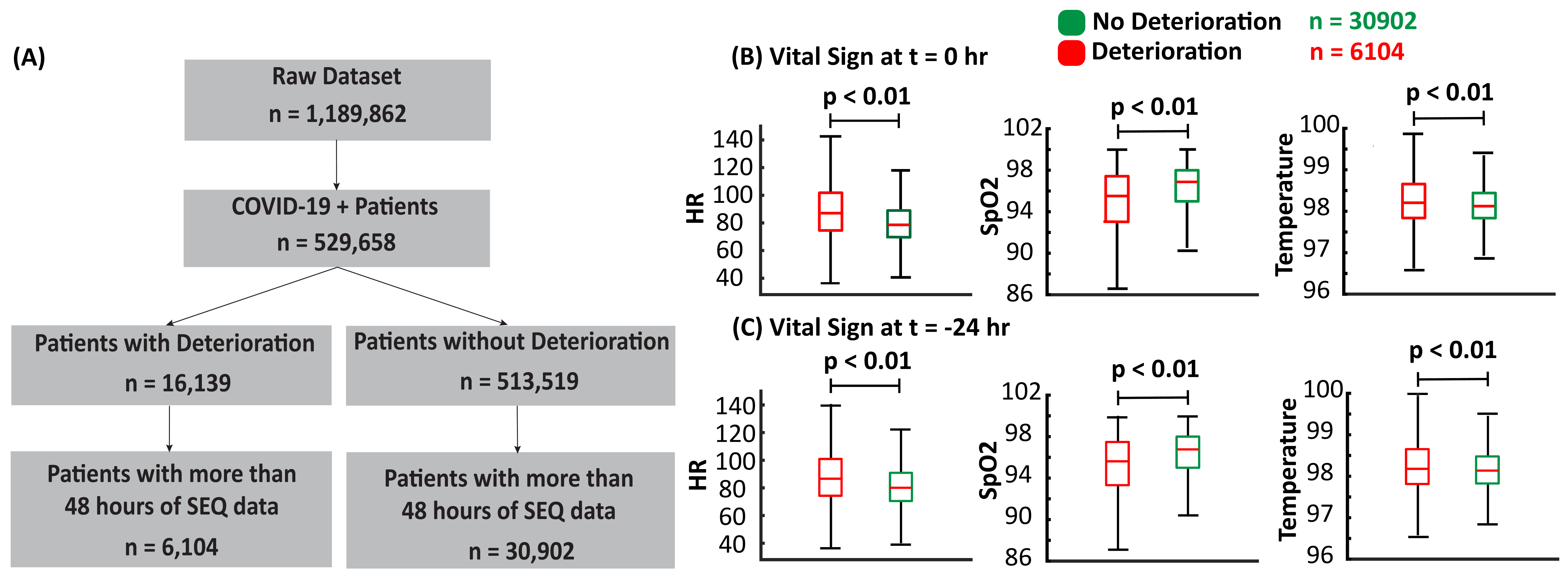}
     \caption{\textbf{Application of data inclusion and exclusion criteria and distribution of vital signs.} \textbf{(A)} In this flowchart, we illustrate the application of the inclusion and exclusion criteria, where \textbf{n} represents the number of patients after each step. \textbf{(B)} The boxplot of the vital signs recorded from the patients at the end of the 24-hour input window (t=0), which corresponds to the prediction time. \textbf{(C)}  The boxplot of the vital signs recorded from the patients at the beginning of the 24-hour input window (t=-24). We observe differences between the two groups (evaluated using the T-test), which motivates the design of the proposed temporal model.}
     \label{fig_3}
    \vspace{-2mm}
\end{figure*}

\begin{figure}[t]
    \includegraphics[width=0.4\textwidth]{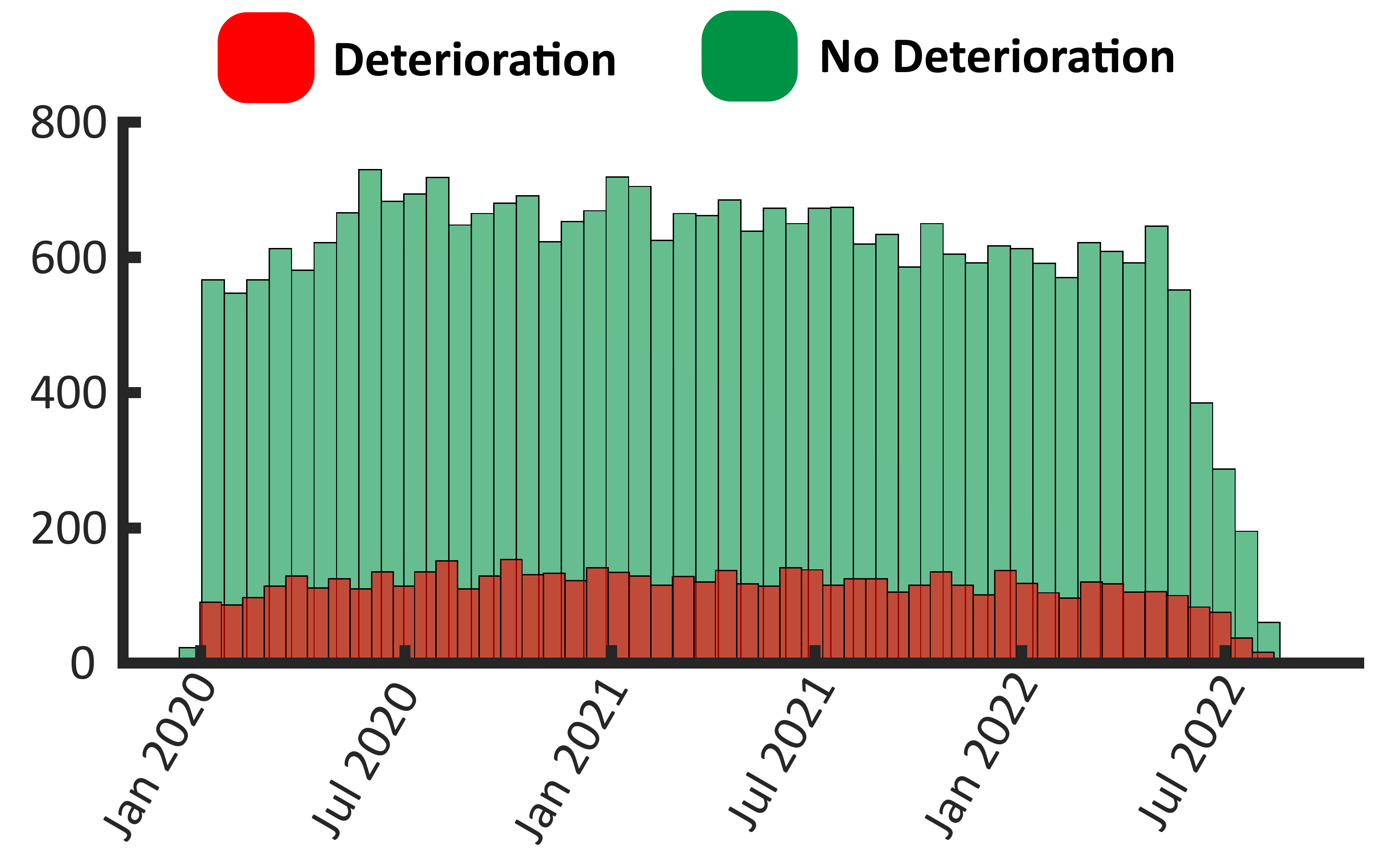}
    \caption{\textbf{Distribution of samples over time.} We show the number of patients who deteriorated vs those who did not deteriorate in our final filtered dataset (n=37,006).}
    \label{fig_2_dist}
    \vspace{-3mm}
\end{figure}

\subsubsection{Three-Phase Training Strategy}
In order to optimize the performance of the proposed network, the training strategy consists of three phases as described below. 

\begin{itemize}
    \item \textbf{Phase 1: \textit{Training of SEQ module}}\\ In the first phase, we train the SEQ module only and freeze all other layers. The output of the FC layer in the SEQ module is connected to another FC layer that computes the prediction.  After this phase of training, the weights are used to pre-initialize SEQ module in the next phase and we remove the second FC layer used to compute predictions during pre-training in the first phase.  
    \item \textbf{Phase 2: \textit{Training of fusion layer}}\\ In the second phase, we compute the representations of the SEQ data after initializing the associated module with the weights obtained in the first phase, and then freeze the SEQ module. We then train the FC layer, FC fusion layer, and the FC output prediction layer using the non-SEQ data. 
    \item \textbf{Phase 3: \textit{End-to-end fine-tuning of SVS-Net}}\\ In the last phase, we initialize the  parameters of the entire network using the weights obtained in the first two phases. The network is trained end-to-end, with the aim of improving the overall network performance. 
\end{itemize}

\subsubsection{Model Training and Evaluation}
To train and evaluate the model, we use a three-fold cross validation process. We randomly divide the entire dataset into three folds so that each fold has the same distribution of positive and negative samples. The final performance results reported are obtained by averaging the validation performance over three  folds.  

We train the model within each fold for 200 epochs using the three-phase training strategy. For the first and second phase, we choose a learning rate of $0.0001$ based on initial experimentation, and a learning rate of $0.00001$ for the final fine-tuning stage. For all phases, we use the ADAM optimizer \cite{74} with $\beta_1 = 0.9$, $\beta_2 = 0.999$, $\epsilon = 10^{-8}$. To avoid over-fitting, we include a patience period of 100 epochs to stop training if the validation loss does not improve. The best model is chosen based on the minimum validation loss. We choose the binary focal cross-entropy loss \cite{69} in order to manage class imbalance in the dataset. We evaluate the performance of the model using three widely used metrics for binary classification:  accuracy (with 0.5 threshold to convert the model predictions into binary), AUROC, and AUPRC.

\section{Results and Discussion}
\subsection{Patient Cohort}
In Figure~\ref{fig_3}(A), we show the application of the inclusion and exclusion criteria. This resulted with 37,006 patient samples, including 6,104 positive samples, and 30,902 negative samples. Table~\ref{tab_2} summarizes the  characteristics of the patients. In Figure~\ref{fig_3}(B), we show the differences in vital-signs of the two cohorts at $t=0$, while in Figure~\ref{fig_3}(C), we show the differences in vital-signs between the two cohorts at $t=-24$. Using t-test statistical analysis, it can be seen that the difference between the two cohorts is statistically significant even as early as 24 hours before deterioration. This motivates the use of the proposed DL to decode the hidden pattern differentiating the two cohorts. In Figure~\ref{fig_2_dist}, we show the distribution of the admitted patients over time.

\subsection{Model Performance}
\begin{figure}[t]
    \includegraphics[width=0.4\textwidth]{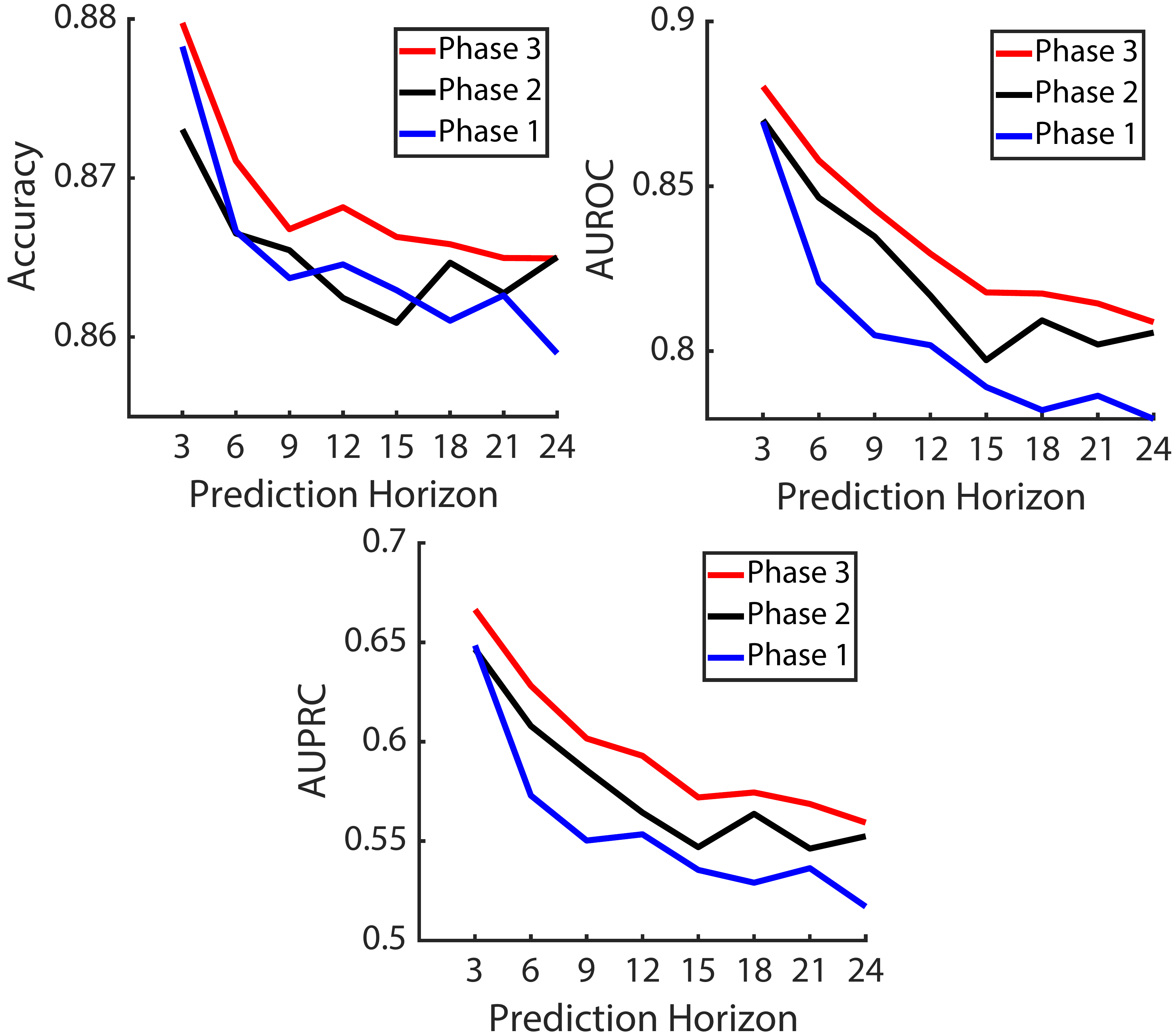}
    \caption{\textbf{Results after each training phase.} Performance results after each phase in the training strategy across all prediction horizons.}
    \label{fig_5}
    \vspace{-3mm}
\end{figure}

The final model performance is summarized in Table~\ref{tab_3} and also shown in Figure~\ref{fig_5} after the three-phase training strategy. The performance at the 24 hours time horizon reaches 86.4\% accuracy, 0.808 AUROC and 0.559 AUPRC. Although the datasets used in these studies are not the same, the results are comparable to those reported previously in~\cite{10} for a subset of this dataset. We also observe that the prediction accuracy improves as the prediction horizon reduces. 

As shown in Figure~\ref{fig_5}, AUROC and AUPRC consistently improve at all prediction horizons after each phase of training, except for the time horizon of three hours, where the AUROC and AUPRC are comparable across phases two and three. This implies that the three-phase training strategy is better suited for longer prediction horizons.

Comparing the results of the performance of the three phases of training, it can be observed that the adopted three-phase training strategy boosts the performance of the model by  forcing the network to extract information initially from the SEQ data and in the end from combination of SEQ and non-SEQ data. The improvement can be seen in  accuracy, AUROC and AUPRC.

\begin{table}[t]
    \centering
    \caption{\textbf{Model performance.} We summarize the performance of the proposed network after the three-phase training stage across all prediction horizons.}
    \begin{tabularx}{\linewidth}{|>{\hsize=1.5\hsize}X |
                              >{\hsize=0.5\hsize}X |
                              >{\hsize=0.5\hsize}X |
                              >{\hsize=0.5\hsize}X |
                              >{\hsize=0.5\hsize}X |
                              >{\hsize=0.5\hsize}X |
                              >{\hsize=0.5\hsize}X |
                              >{\hsize=0.5\hsize}X |
                              >{\hsize=0.5\hsize}X|}
      \hline
      \textbf{Prediction horizon (hours)} & \textbf{3} & \textbf{6} & \textbf{9} & \textbf{12} & \textbf{15} & \textbf{18} & \textbf{21} & \textbf{24} \\ 
      \hline
      \textbf{Accuracy} & 0.879  &  0.871  &  0.866   & 0.868  &  0.866  &  0.865  &  0.865  &  0.864\\
      \hline
      \textbf{AUROC} & 0.880  &  0.857   & 0.843   & 0.829   & 0.817  &  0.817  &  0.814  &  0.808\\
      \hline
      \textbf{AUPRC} &  0.666  &  0.628  &  0.601 &   0.592  &  0.572  &  0.574  &  0.568  &  0.559\\
      \hline
    \end{tabularx}
    
    \label{tab_3}
\end{table}

\subsection{Ablation Studies}

In order to understand  the impact of our design choices within the model architecture, we compare our model to two other networks. The first model, referred to as \textit{Memory-Less Vital Sign Network (MLVS-Net)}, processes the non-SEQ data and only the last set of vital-signs collected from the patient, ignoring any sequential information. Hence, instead of using a dilated LSTM, we process the vital-sign data (3 features) with a MLP consisting of two FC layers with output dimensionality of 16 each. The computed representation of the MLP is then concatenated with the representation of the non-SEQ data. We train the model in a similar fashion using the three-phase training strategy, and we freeze the weights of the MLP network in phase two. The second model, referred to as the \textit{non-Sequential Health Status Network (nSHS-Net)}, only considers the non-SEQ data. Hence, the output of the FC layer is processed by a second FC layers to generate the prediction.

\begin{figure}[t]
    \includegraphics[width=0.4\textwidth]{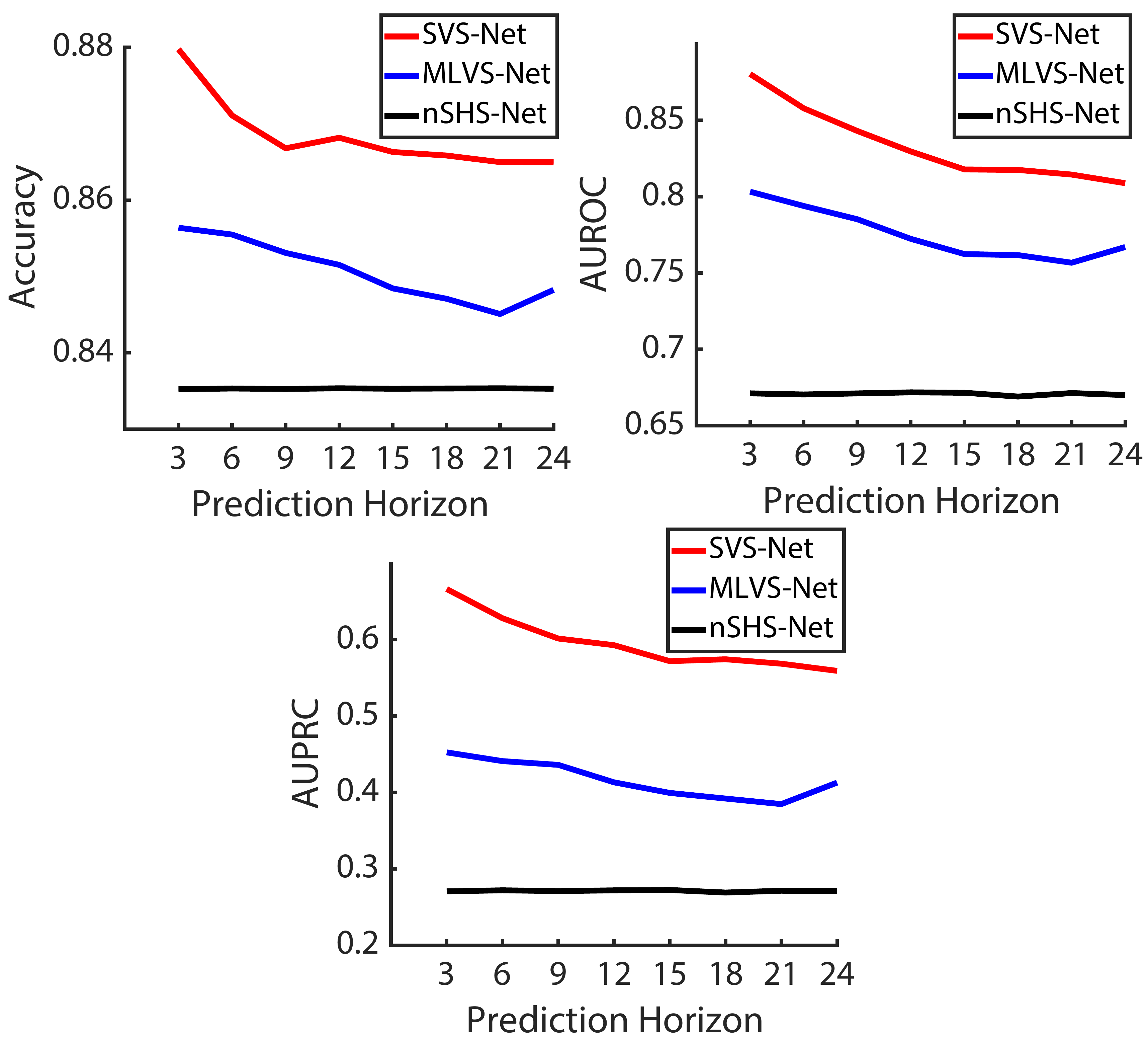}
    \caption{\textbf{Ablation study results.} Performance results for each of SVS-Net (non-SEQ data and SEQ vital sign data), MLVS-Net (non-SEQ data and single set of vital-signs), and nSHS-Net (non-SEQ only).}
    \label{fig_6}
    \vspace{-5mm}
\end{figure}

We compare the three models in Figure~\ref{fig_6}. First, we observe that \textit{nSHS-Net}  performs the worst, which implies that the incorporation of vital signs is crucial for the model prediction. When comparing, \textit{MLVS-Net} and \textit{SVS-Net}, we observe a better performance with the latter across all prediction horizons and evaluation metrics. This implies that the incorporation of sequential information can significantly improve the capability of the model in predicting deterioration, relative to using a single measurement of vital signs. The numerical results of Figure~\ref{fig_6} are summarized in Appendix I, Table IV.

\begin{figure*}[t!]
\centering

     \includegraphics[width=0.9\textwidth]{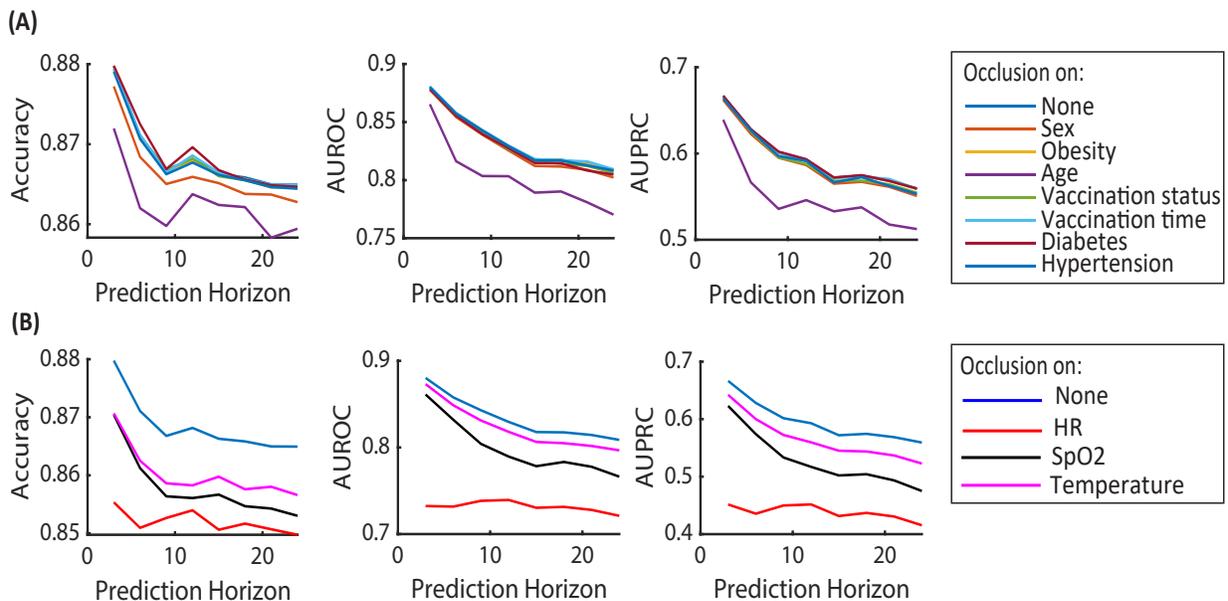}
     \caption{\textbf{Results of occlusion analysis to understand the importance of the input features.} \textbf{(A)}: Occlusion analysis on the clinical and comorbidity characteristics data. Occlusion of Age  
         decreases the model performance (across all three evaluation metrics) more significantly than others when occluded. \textbf{(B)}: Occlusion analysis on SEQ vital sign data. it can be observed that the HR contributes more significantly to the model performance than the SpO2 and temperature. 
     }
     \label{fig_7}
\end{figure*}

\subsection{Occlusion Analysis}
In order to assess the influence of each input variable, we conduct an occlusion analysis on the final optimized model. In particular, we occlude one feature at a time (by setting the corresponding values to zero) and evaluate the performance on the validation set. The greater the reduction in the performance metrics upon feature occlusion, the more important the feature is for the prediction.

The results of this analysis are shown in Figure~\ref{fig_7}. As shown in Figure~\ref{fig_7}(A), amongst the non-SEQ features, we observe that age plays the most  significant role in the model's performance. The  other CCC features have less of an impact once occluded and do not have consistent trends across the different performance metrics and prediction horizons. It should be noted that due to the correlation between the various features, some features may be relevant to the prediction task, yet are not considered to be important by this occlusion analysis. 

For example if one CCC feature affects the variations in the vital signs over time, then the corresponding effect would be captured by the model. In this case, the occlusion analysis may show that this CCC feature is not important.  However, when a feature shows low sensitivity through the occlusion study, it means that the need for that feature to be given  to the model as an ``independent'' input is not significant. Among the SEQ features, we observe heart rate as the most important feature, followed by SpO2, and then temperature. 

It should be highlighted that in Fig. 7, we have shown that the vital signs are more important than the clinical and comorbidity features, and SEQ vital signs bring additional information than the  current vital signs only. The observation here matches the aforementioned analysis. Numerical results of Fig. 7 is given in Appendix I, Table V.

\section{Conclusion}

This study is motivated by the high availability of personal medical devices, such as wearable systems (e.g., smart watches), that can record time-series medical data, and the prospects of using such devices in the context of telehealth for the prediction of deterioration. In summary, we propose, develop, and evaluate a deterioration prediction model using a large dataset (n=37,006) collected at  NYU Langone Health during the period of January 2020 to September 2022. The model achieves an AUROC of 0.808-0.880 in 3 to 24 hours prediction horizons. 

Our study has several strengths. First, the model uses a minimal input feature set consisting of time-series of three routine vital signs, i.e., SpO2, heart rate, and temperature. The model is also provided with basic patient information, including sex, age,  vaccination status, vaccination date, presence of obesity, hypertension, and diabetes, which can be easily collected. Compared to previous work \cite{10} that achieved 0.765 AUORC,  our model achieves a comparable performance.  However, that model was trained and evaluated using a different data modalities.  In order to assess the significance of the various clinical and comorbidity features and the importance of using time-series vital-signs data, we performed a sensitivity analysis through an occlusion experiment as well as an ablation study. The results showed the importance of modeling the temporal variations of vital signs and the possibility of achieving high prognosis accuracy without the need for sophisticated medical imaging. Finally, the proposed framework is scalable as it can be extended for other prediction horizon ranges and using windows of different lengths.

The proposed work also has limitations. First, our input windows are limited to a size of 24 hours. In future work, we are interested in varying the length of the input feature windows and investigate the impact on performance, with the goal of reducing computational complexity if similar accuracy results can be obtained with shorter windows. Second, we do not perform any external or prospective validation of the model due to lack of access to similar datasets. Finally, we believe that the final results can be improved via hyperparameter tuning, including the learning rate, and this is an area of future work. 

To conclude, this study highlights the feasibility of an accessible and scalable model to help assist the medical workforce in decision-making. The versatility of the proposed model is of importance, as the data types used for training and evaluating the model can be easily acquired from patients using wearable sensors and a few clinical data features that can be self-reported.

\section{Acknowledgement}

$\bullet$ { Funding:} This material is based upon work supported by the National Science Foundation (Award \# 2031594).

     $\bullet$ { Conflict of Interest:} S. Farokh Atashzar and Yao Wang are inventors of “Smart Wearable IOT Device for Health Tracking, Contact Tracing and Prediction of Health Deterioration” which is licensed by Tactile Robotics, Ltd., Canada. 

\bibliographystyle{IEEEtran}
\bibliography{Ref}

\appendices

\section*{Appendix I}

Here we provide numerical results with respect to Fig~\ref{fig_6} in Table~\ref{tab_5}, and results with respect to Fig~\ref{fig_7} in Table~\ref{tab_6}.

\begin{table}[h!]
    \centering
    \caption{Numerical comparison of SVS-Net, MLVS-Net, and nSHS-Net for 3-24 hours of prediction horizons}
    \begin{tabularx}{\linewidth}{|>{\hsize=1.5\hsize}X |
                              >{\hsize=0.5\hsize}X |
                              >{\hsize=0.5\hsize}X |
                              >{\hsize=0.5\hsize}X |
                              >{\hsize=0.5\hsize}X |
                              >{\hsize=0.5\hsize}X |
                              >{\hsize=0.5\hsize}X |
                              >{\hsize=0.5\hsize}X |
                              >{\hsize=0.5\hsize}X|}
      \hline
      \textbf{Prediction Horizon (hours)} & \textbf{3} & \textbf{6} & \textbf{9} & \textbf{12} & \textbf{15} & \textbf{18} & \textbf{21} & \textbf{24} \\ 
      \hline
      \multicolumn{9}{|c|}{\textbf{Accuracy}} \\
      \hline
      \textbf{SVS-Net} & \textbf{0.879}  &  \textbf{0.871}  &  \textbf{0.866}  &  \textbf{0.868}  &  \textbf{0.866}   & \textbf{0.865}  &  \textbf{0.865} &  \textbf{0.864}\\
      \hline
      \textbf{MLVS-Net} & 0.856 &  0.855 & 0.853 & 0.851 & 0.848 & 0.847 & 0.845 & 0.848\\
      \hline
      \textbf{nSHS-Net} & 0.835  &  0.835  &  0.835   & 0.835 &   0.835  &  0.835  &  0.835  &  0.835\\
      \hline
      \multicolumn{9}{|c|}{\textbf{AUROC}} \\
      \hline
      \textbf{SVS-Net} &\textbf{0.880}  &  \textbf{0.857}  &  \textbf{0.843}  &  \textbf{0.829}   &\textbf{0.817}  & \textbf{0.817} &  \textbf{0.814} &   \textbf{0.808}\\
      \hline
      \textbf{MLVS-Net} & 0.803 &   0.793 &   0.785  &  0.772  &  0.762 & 0.761  &  0.756   & 0.767\\
      \hline
      \textbf{nSHS-Net} & 0.671  &  0.670   & 0.671   & 0.671   & 0.671  &  0.669 &   0.671  &  0.670\\
      \hline
      \multicolumn{9}{|c|}{\textbf{AUPRC}} \\
      \hline
      \textbf{SVS-Net} & \textbf{0.666}  &  \textbf{0.628}  &  \textbf{0.601}   & \textbf{0.592}  &  \textbf{0.572}  &  \textbf{0.574} &   \textbf{0.568}   & \textbf{0.559}\\
      \hline
      \textbf{MLVS-Net} & 0.496 &   0.484  &  0.474  &  0.458  &  0.437  &  0.432  &  0.426 &   0.453\\
      \hline
      \textbf{nSHS-Net} &0.270  &  0.271  &  0.270  &  0.271  &  0.272   & 0.269  &  0.271 &   0.271\\
      \hline
    \end{tabularx}
    
    \label{tab_5}
\end{table}

\begin{table}[h!]
    \centering
    \caption{Numerical comparison of all the occlusions on clinical and comorbidity characteristics and SEQ vital sign data for 3-24 hours of prediction horizons}
    \begin{tabularx}{\linewidth}{|>{\hsize=1.7\hsize}X |
                              >{\hsize=0.5\hsize}X |
                              >{\hsize=0.5\hsize}X |
                              >{\hsize=0.5\hsize}X |
                              >{\hsize=0.5\hsize}X |
                              >{\hsize=0.5\hsize}X |
                              >{\hsize=0.5\hsize}X |
                              >{\hsize=0.5\hsize}X |
                              >{\hsize=0.5\hsize}X|}
      \hline
      \textbf{Prediction Horizon (hours)} & \textbf{3} & \textbf{6} & \textbf{9} & \textbf{12} & \textbf{15} & \textbf{18} & \textbf{21} & \textbf{24} \\ 
      \hline
      \multicolumn{9}{|c|}{\textbf{Accuracy}} \\
      \hline
      \textbf{None} & 0.879  &  0.871  &  0.866  &  0.868  &  0.866   & 0.865  &  0.865 & 0.864\\
      \hline
      \textbf{Sex} & 0.877  &  0.868  &  0.865  &  0.865  &  0.865   & 0.863 &   0.863  &  0.862\\
      \hline
      \textbf{Obesity} & 0.879  &  0.870  &  0.866  &  0.867  &  0.866  &  0.865 &   0.864  &  0.864\\
      \hline
      \textbf{Age} & 0.872  &  0.862  &  0.859  &  0.863  &  0.862  &  0.862 &   0.858 &   0.859\\
      \hline
      \textbf{Diabetes} & 0.879 &   0.872  &  0.866  &  0.869  &  0.866  &  0.865  &  0.864  &  0.864\\
      \hline
      \textbf{Hypertension} &0.879 &   0.870 &   0.866   & 0.867 &   0.866  &  0.865 &   0.864 &    0.864\\
      \hline
      \textbf{Vac. Time} & 0.879 &   0.871  &  0.866 &   0.868  &  0.866  &  0.865  &  0.864&    0.864\\
      \hline
      \textbf{Vac. Status} & 0.879 &   0.871  &  0.866 &   0.868   & 0.866   & 0.865 &   0.864&    0.864\\
      \hline
      \textbf{HR} & 0.855 &   0.850   & 0.852  &  0.854 &  0.850 &   0.851  &  0.850  &  0.849\\
      \hline
      \textbf{SpO2} & 0.870  &  0.861  &  0.856   & 0.856  &  0.856  &  0.854 &   0.854 &   0.853\\
      \hline
      \textbf{Temperature} & 0.870  &  0.862  &  0.858  &  0.858  &  0.859  &  0.857  &  0.858 &   0.856\\
      \hline
      
      \multicolumn{9}{|c|}{\textbf{AUROC}} \\
      \hline
      \textbf{None} & 0.880 & 0.858 & 0.843 & 0.830 & 0.818 & 0.817 & 0.814 & 0.809\\
      \hline
      \textbf{Sex} & 0.878 & 0.854 & 0.839 & 0.826 & 0.812 & 0.812 & 0.809 & 0.802\\
      \hline
      \textbf{Obesity} & 0.880 & 0.858 & 0.843 & 0.829 & 0.817 & 0.817 & 0.814 & 0.808\\
      \hline
      \textbf{Age} & 0.865 & 0.816 & 0.804 & 0.803 & 0.789 & 0.790 & 0.781 & 0.770\\
      \hline
      \textbf{Diabetes} & 0.878 & 0.855 & 0.840 & 0.827 & 0.815 & 0.815 & 0.808 & 0.805 \\
      \hline
      \textbf{Hypertension} & 0.880 & 0.857 & 0.843 & 0.829 & 0.817 & 0.817 & 0.813 & 0.808\\
      \hline
      \textbf{Vac. Time} & 0.881 & 0.857 & 0.843 & 0.829 & 0.818 & 0.817 & 0.816 & 0.809\\
      \hline
      \textbf{Vac. Status} & 0.880 & 0.857 & 0.840 & 0.826 & 0.816 & 0.814 & 0.812 & 0.806\\
      \hline
      \textbf{HR} & 0.732 & 0.732 & 0.738 & 0.739 & 0.730 & 0.731 & 0.728 & 0.721\\
      \hline
      \textbf{SpO2} & 0.861 & 0.832 & 0.804 & 0.790 & 0.778 & 0.783 & 0.778 & 0.766\\
      \hline
      \textbf{Temperature} & 0.873 & 0.849 & 0.831 & 0.818 & 0.806 & 0.805 & 0.802 & 0.797\\
      \hline
      
      \multicolumn{9}{|c|}{\textbf{AUPRC}} \\
      \hline
      \textbf{None} & 0.666 & 0.628 & 0.602 & 0.593 & 0.572 & 0.574 & 0.569 & 0.559\\
      \hline
      \textbf{Sex} & 0.662 & 0.623 & 0.595 & 0.587 & 0.565 & 0.567 & 0.561 & 0.551\\
      \hline
      \textbf{Obesity} & 0.666 & 0.627 & 0.601 & 0.592 & 0.571 & 0.574 & 0.568 & 0.559\\
      \hline
      \textbf{Age} & 0.639 & 0.567 & 0.536 & 0.546 & 0.533 & 0.538 & 0.518 & 0.512\\
      \hline
      \textbf{Diabetes} & 0.664 & 0.627 & 0.597 & 0.591 & 0.567 & 0.572 & 0.563 & 0.553 \\
      \hline
      \textbf{Hypertension} & 0.667 & 0.628 & 0.602 & 0.593 & 0.572 & 0.575 & 0.568 & 0.559\\
      \hline
      \textbf{Vac. Time} & 0.666 & 0.628 & 0.601 & 0.591 & 0.572 & 0.573 & 0.570 & 0.559\\
      \hline
      \textbf{Vac. Status} & 0.664 & 0.625 & 0.596 & 0.588 & 0.568 & 0.569 & 0.564 & 0.555\\
      \hline
      \textbf{HR} & 0.452 & 0.436 & 0.450 & 0.452 & 0.432 & 0.437 & 0.431 & 0.415\\
      \hline
      \textbf{SpO2} & 0.623 & 0.574 & 0.533 & 0.517 & 0.502 & 0.504 & 0.493 & 0.475\\
      \hline
      \textbf{Temperature} & 0.642 & 0.600 & 0.572 & 0.560 & 0.545 & 0.544 & 0.537 & 0.523\\
      \hline
      
    \end{tabularx}\vspace{9.5cm}
    
    \label{tab_6}
\end{table}

\end{document}